\documentclass[sigconf]{acmart}
\setcopyright{none}
\renewcommand\footnotetextcopyrightpermission[1]{} 

\AtBeginDocument{%
  \providecommand\BibTeX{{%
    \normalfont B\kern-0.5em{\scshape i\kern-0.25em b}\kern-0.8em\TeX}}}

\acmBooktitle{Woodstock '18: ACM Symposium on Neural Gaze Detection,
  June 03--05, 2018, Woodstock, NY}
\acmPrice{15.00}
\acmISBN{978-1-4503-XXXX-X/18/06}

\acmSubmissionID{2627}

\usepackage{times}
\usepackage{epsfig}
\usepackage{graphicx}
\usepackage{bm}
\usepackage{algorithm, algpseudocode}
\usepackage{multirow}
\usepackage{caption}
\usepackage{comment}
\usepackage{color}
\definecolor{citecolor}{RGB}{119,185,0} 

\usepackage{pifont}

\usepackage{amssymb}
\usepackage{amsmath}
\usepackage{animate}
\usepackage{graphicx}

\usepackage{booktabs}
\usepackage{multirow}
\usepackage{makecell}
\usepackage{adjustbox}

\newlength\savewidth

\def\eg{\emph{e.g.}} 
\def\ie{\emph{i.e.}} 
 
\def\etc{\emph{etc}} 

\begin{document}
\title{MRTNet: Multi-Resolution Temporal Network for Video Sentence Grounding}

\author{Wei Ji}
\affiliation{%
 \institution{National University of Singapore}
 \country{Singapore}}
 \email{weiji0523@gmail.com}
 
 \author{Long Chen}
\affiliation{%
 \institution{Columbia University}
 \country{USA}}
 \email{zjuchenlong@gmail.com}
 
  \author{Yinwei Wei}
\affiliation{%
 \institution{National University of Singapore}
 \country{Singapore}}
 \email{weiyinwei@hotmail.com}
 
   \author{Yiming Wu}
\affiliation{%
 \institution{ The University of Sydney}
 \country{Australia}}
 \email{yimingwu0@gmail.com}

\author{Tat-Seng Chua}
\affiliation{%
 \institution{National University of Singapore}
 \country{Singapore}}
 \email{dcscts@nus.edu.sg}


\begin{abstract}
  Given an untrimmed video and natural language query, video sentence grounding aims to localize the target temporal moment in the video. Existing methods mainly tackle this task by matching and aligning semantics of the descriptive sentence and video segments on a single temporal resolution, while neglecting the temporal consistency of video content in different resolutions. In this work, we propose a novel multi-resolution temporal video sentence grounding network: MRTNet, which consists of a multi-modal feature encoder, a Multi-Resolution Temporal (MRT) module, and a predictor module. MRT module is an encoder-decoder network, and output features in the decoder part are in conjunction with Transformers to predict the final start and end timestamps. Particularly, our MRT module is hot-pluggable, which means it can be seamlessly incorporated into any anchor-free models. Besides, we utilize a loss function to supervise cross-modal features in MRT module for more accurate grounding in three scales: frame-level, clip-level and sequence-level. Extensive experiments on three prevalent datasets have shown the effectiveness of MRTNet.
\end{abstract}


\maketitle

\section{Introduction}
Video sentence grounding aims to localize the temporal moments from the untrimmed video corresponding to the given descriptive language query. As an important visual-language task, video sentence grounding involves both computer vision (CV) and natural language processing (NLP) techniques, and is beneficial to numerous down-stream video understanding tasks, such as video dialog~\cite{chu2021end}, video relationship detection~\cite{shang2017video,shang2021video,ji2021vidvrd}, and video question answering~\cite{zhong2022video,xiao2022video,li2022invariant}, \etc.
Clearly, the cross-modal semantic consistency is essential for video sentence grounding.

\begin{figure}[!tbp]
    \centering
    \includegraphics[width=1\linewidth]{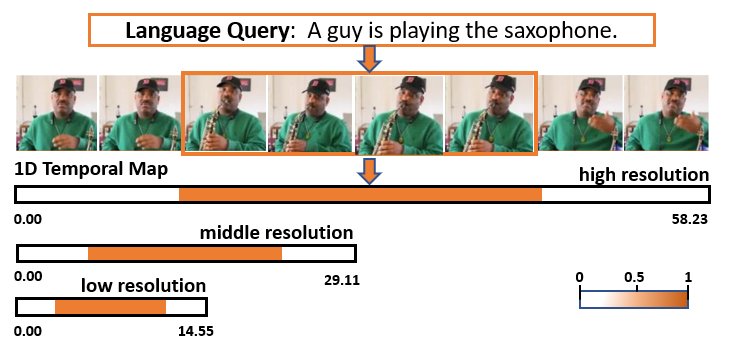}
    \caption{An illustration of localizing a temporal moment in an untrimmed video by a language query. The 1D temporal map shows the response area in the whole video sequence, according to the language query. As shown above, if video is compressed with different sampling rates, these 1D temporal maps will display in high, middle, and low resolutions, while the relative locations of target regions stay unchanged.}
    \label{fig:introduction}
\end{figure}

Prior works primarily treat video sentence grounding as a ranking problem, \ie, applying a multi-modal matching architecture to find the best matching moment for the given language query, which can be treated as proposal-based methods. Recently, some works explore the cross-modal matching between video and query, and then directly regress the temporal locations of target moment, which is classified as proposal-free methods. Commonly, the structure of current proposal-free video language grounding methods mainly includes these modules: visual and language encoders module, multi-modal attention module, and predictor module. The results calculated by the predictor module heavily rely on the multi-modal features with attention.
However, these methods neglect the temporal consistency of video data in multi-resolutions, which is beneficial for better video feature representation.

In this paper, we try to deal with the video sentence grounding task from another perspective: In the 1D temporal dimension, the target range in video can be treated as the foreground region, and the rest of the video is viewed as the background. The essence of video sentence grounding is to localize the maximum response area in 1D temporal range according to the descriptive query. As we known, attention map can also be treated as the maximum response area in the temporal dimension. Hence, context information in the whole video is crucial for locating the accurate boundaries of temporal segments.

On the basis of the previous network design, we propose a plug-and-play module to constrains the temporal consistency of video in different resolutions.
In the module design, we aim to quickly localize the relative location of the target range in low resolution of the video by downsampling, and then refine the accurate boundaries in the decoder part when upsampling to the original resolution.
Inspired by the observations above, we propose a Multi-Resolution Temporal Network (MRTNet) for the video sentence grounding task to simultaneously address all the above-mentioned limitations. It mainly consists of a multi-model feature encoder, a multi-resolution temporal (MRT) module, and a predictor module. Our MRT model is hot-pluggable, and can be seamlessly incorporated into any stronger proposal-free or proposal-based video sentence grounding network. Besides, we utilize a vision-and-linguistic Transformer to encode the multi-modal features with attention. Moreover, we propose a novel loss function that fuses Cross Entropy (CE) loss, Structural Similarity Index Measure (SSIM) loss, and Intersection over Union (IoU) loss to supervise the training process of accurate video sentence grounding in three levels: frame-level, clip-level, and sequence-level. We demonstrate the effectiveness of the proposed MRTNet on multiple benchmarks. From the experimental results and ablation study, we observe consistent improvements across extensive ablations and settings. 

In summary, our proposed MRTNet has the following advantages:

\begin{itemize}
    \item We propose a novel multi-resolution temporal video sentence grounding network: MRTNet, which mainly consists of a multi-model feature encoder, a multi-resolution temporal (MRT) module, and a predictor module. We use a multi-modal transformer to encode the visual and linguistic features. Our MRT model is hot-pluggable, and can be seamlessly incorporated into any stronger proposal-free or proposal-based video sentence grounding network.

    \item We propose a loss function that fuses CE loss, SSIM loss, and IoU loss to supervise the training process of accurate video sentence grounding in three levels: frame-level, clip-level, and sequence-level.

    \item Experimental results show that our MRT module can effectively improve the performance of different baseline methods in terms of popular evaluation metrics.
\end{itemize}

\begin{figure*}
    \centering
    \includegraphics[width=\linewidth,height=7cm]{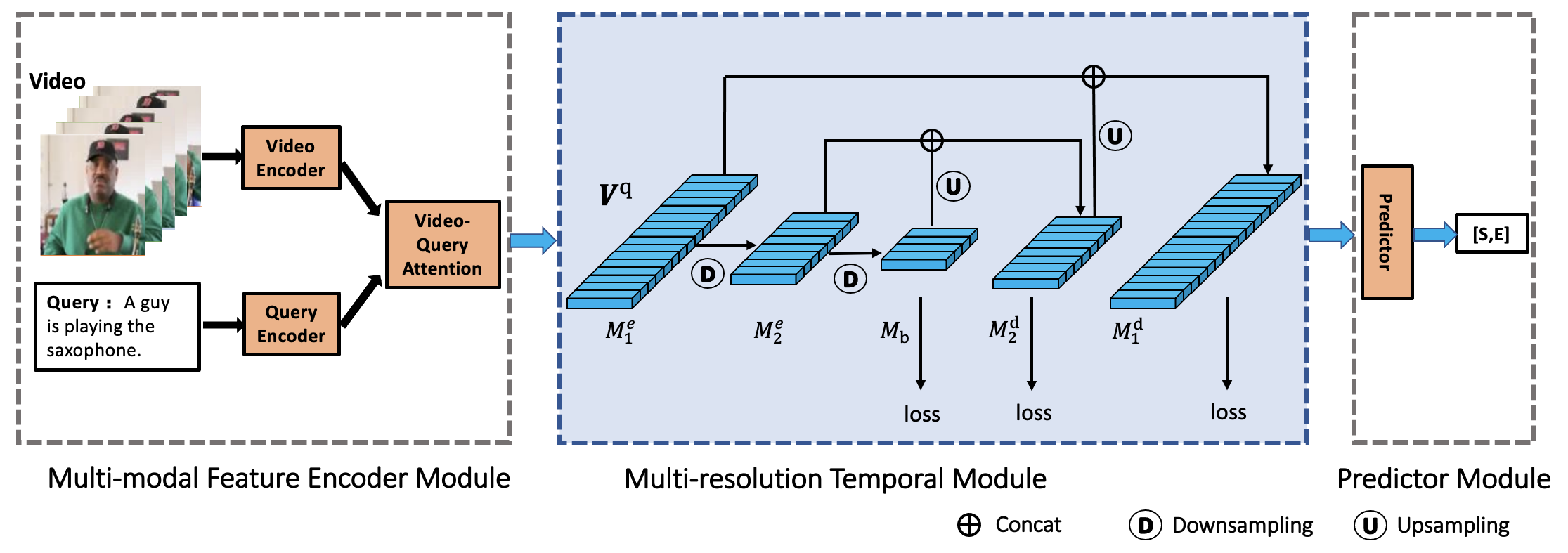}
    \caption{The overview architecture of the MRTNet model, which consists of three parts, including Multi-modal Feature Encoder Module, Multi-resolution Temporal Module, and Predictor Module. Taking the untrimmed video and language query as input, the Feature Encoder Module first extracts the video and query feature by Video Encoder and Query Encoder separately. Then cross-modal features are fused by video-query attention mechanism. Taking the cross-modal features as input, multi-resolution temporal module use an encoder-decoder architecture with downsampling and upsampling, and regularizes the context temporal information from different resolutions. Finally, the temporal boundary results (start and end timestamps) are predicted according to the temporal context-aware features.} 
    \label{fig:architecture}
\end{figure*}

\section{Related Work}

\subsection{Video Sentence Grounding}
Video sentence grounding is defined as retrieving video segments corresponding to language queries~\cite{anne2017localizing,xiao2021boundary,liu2020jointly,liu2021context}. Most proposed methods~\cite{hendricks2018localizing,liu2018attentive} treat this task as a ranking problem. They first generate a certain amount of proposals of video segments, and then match the visual content of proposals and linguistic information of language queries. After ranking segment proposals according to the multi-modal matching, the best matching moment for the query can be selected. Similar to object detection, which also generates proposals first and then classifies the content in the bounding box, mining appropriate numbers of positive and negative samples are important for model performance. Thus, many methods always densely sample video moment proposals to achieve good performance, which leads to large computation costs.

To overcome the above-mentioned drawbacks, \cite{yuan2019find} propose a proposal-free method, which utilizes a Bi-LSTM to encode visual and sentence features, and a co-attention interaction module to fuse multi-modal features. The model treats the video as a whole and directly predicts the temporal coordinates according to the sentence queries. \cite{lu2019debug} propose a dense bottom-up framework, which treats all frames corresponding to the language query as foreground, and then regresses the unique distances of each frame in the foreground to bi-directional ground-truth boundaries, finally fuses appropriate temporal candidates as final result. 
To mine the relationship of sentence semantics with diverse video contents, Yuan et al. ~\cite{yuan2019semantic} also propose a semantic conditioned dynamic modulation, which leverages sentence semantic information to modulate the temporal convolution processes in a hierarchical temporal convolutional network, and establishes a precise matching relationship between sentence and video. 
Recently, \cite{zhang2020learning} propose a 2D temporal map to model the temporal relations of different moments with variant length, in which the two dimensions indicate the start and end timestamps, respectively.
Apart from these top-down and bottom-up methods, there are also works~\cite{wang2019language}~\cite{he2019read} which adopt reinforcement learning to make decisions on the action space of candidate segments, such as the start or end boundaries move left or right, corresponding to the language query matching result.

Among previous works, most similar to ours is DRN ~\cite{zeng2020dense}. 
They propose a dense regression network with three different heads to generate accurate boxes. However, directly training three head branches is difficult and unstable, the authors propose a three-step strategy to train the proposed DRN~\cite{zeng2020dense}.
Different from generating candidate boxes in different scales, we propose a multi-resolution temporal network to further leverage the benefit of 1D temporal attention to better model the context information in different resolutions, which improves a lot for the accurate temporal region in videos. 
Besides, to obtain accurate attention information, we utilize a loss function to regularize the attention map from three different levels respectively. Furthermore, our proposed module can be plugged into other anchor-free baselines and can be easily trained in an end-to-end manner.

\subsection{Multi-modal Transformer} 
Transformer~\cite{vaswani2017attention} is first proposed in natural language processing community, which consists of the encoder and decoder with self-attention to deal with the neural machine translation problem. 
Then, transformer-based networks are proposed to deal with various NLP tasks, such as language modeling~\cite{dai2018transformer}, word sense disambiguation~\cite{tang2018self}, and learning sentence representations~\cite{devlin2018bert}.
Due to the great success of Transformer in NLP tasks, its applications to vision tasks have received increasing attention~\cite{dosovitskiy2020image}.
By splitting the image into fixed-size patches, an image can be treated as a sequence of vectors, which is the same as a sentence. Visual Transformer can be applied in a serious of CV tasks, such as image classification~\cite{wang2021crossformer}, semantic segmentation~\cite{wang2021pyramid}, \etc.

As for the multimedia area, several works have tried to utilize Transformer architecture to deal with the multi-modal feature interaction.
For example, \cite{gabeur2020multi} propose a multi-modal transformer to jointly encode the different modalities in video, so as to retrieve the corresponding video according to language and audio. \cite{yu2019multimodal} propose a Multi-modal Transformer model for image captioning, which simultaneously captures intra-modal and inter-modal interactions in a unified attention block. In this paper, we propose to use Transformers to encode the multi-modal features and predict the start and end timestamps.

\section{Approach}
In this paper, we propose an MRTNet model to deal with the video sentence grounding task. To sum up, MRTNet mainly consists of three parts: multi-model feature encoder module, multi-resolution temporal module, and a predictor module. The whole architecture of MRTNet is illustrated in Figure~\ref{fig:architecture}. In this section, we first define the problem formulation and describe the multi-model feature encoder module. Then, we present the details of the Multi-Resolution Temporal (MRT) module. Finally, we calculate the start and end timestamps in the predictor module and show the training and inference processes of MRTNet in detail.

\subsection{Problem Formulation}\label{Problem Formulation}

Given an untrimmed video $V={f_t}_{t=1}^{T}$ and the language query $Q={q_j}_{j=1}^{M}$, where $T$ and $M$ are the numbers of frames and words, respectively, our goal is to predict the start and end timestamp ($\tau^{s}$, $\tau^{e}$) in the video corresponding to query $Q$. For each video $V$, we extract its visual features $\bm{V}={\bm{v}_i}_{i=1}^{n}$ with a pretrained 3D ConvNet, where $n$ is the length of extracted features, each feature $\bm{v}_i$ here is a video feature vector. For each query $Q$, we initialize the word features
${\bm{Q}=\bm{q}_j}_{j=1}^{m}$ by using GloVe embeddings.

\subsection{Multi-modal Feature Encoder Module}\label{ssec:feature}

 For the visual feature $\bm{V}\in\mathbb{R}^{n\times d_{v}}$ and linguistic feature $\bm{Q}\in\mathbb{R}^{m\times d_{q}}$, We first project them into the same dimension $d$ by two linear layers. Then we adopt the same embedding encoder layer as QANet ~\cite{yu2018fast} to build the visual and linguistic feature encoder, which consists of four convolution layers, a multi-head attention layer~\cite{vaswani2017attention}, a layer normalization layer, and a feed-forward layer with the residual connection. The encoded visual features and query features are as follows:
\begin{equation}
\begin{aligned}
    \bm{\widetilde{V}} & = \mathtt{VisualEncoder}(\bm{VW^{v}}),  \\
    \bm{\widetilde{Q}} & = \mathtt{QueryEncoder}(\bm{QW^{q}})
\end{aligned}
\end{equation}
where $\bm{W^{v}}$ and $\bm{W^{q}}$ are projection matrices to keep the dimension consistency between two modalities.
The parameters of feature encoder are shared by two modal features.

\noindent\textbf{Multi-modal Attention.}\label{ssec:attention}
With the encoded visual and query features, we calculate the similarity of two modal features and fuse the multi-modal features by a multi-modal attention mechanism.
Similar to~\cite{lu2019debug}, we first calculate the similarity scores, $\mathcal{S}\in\mathbb{R}^{n\times m}$, between each visual feature and query feature. Then the attention weights of visual-to-query ($\mathcal{A}$) and query-to-visual ($\mathcal{B}$) are computed as:
\begin{equation}
\begin{aligned}
    \mathcal{A}=\mathcal{S}_{r}\cdot\bm{\widetilde{Q}}\in\mathbb{R}^{n\times d}, \\
    \mathcal{B}=\mathcal{S}_{r}\cdot\mathcal{S}_{c}^{T}\cdot\bm{\widetilde{V}}\in\mathbb{R}^{n\times d},
\end{aligned}
\end{equation}
where $\mathcal{S}_{r}$ and $\mathcal{S}_{c}$ are the row-wise and column-wise normalization of $\mathcal{S}$ by softmax operation, respectively. Finally, the output of visual-query attention is written as:
\begin{equation}
    \bm{V}^{q}=\mathtt{FFN}\big([\bm{\widetilde{V}};\mathcal{A};\bm{\widetilde{V}}\odot\mathcal{A};\bm{\widetilde{V}}\odot\mathcal{B}]\big),
\end{equation}
where $\bm{V}^{q}\in\mathbb{R}^{n\times d}$; $\mathtt{FFN}$ is a single feed-forward layer; $\odot$ denotes element-wise multiplication. $\bm{V}^{q}$ is the fused multi-modal semantic features with visual and query attention.

\subsection{Multi-Resolution Temporal Module}\label{MRT Network}
The multi-modal features $\bm{V}^{q}$ we obtain above are in the same length as the video feature $\bm{V}$. However, the temporal consistency of video sequence in different resolutions lacks of exploration.
Inspired by the U-Net~\cite{ronneberger2015u}, we design a multi-resolution temporal module in an encoder-decoder fashion. This encoder-decoder design is able to capture contexts globally in low resolution and refine accurate temporal boundary information in high resolution at the same time. Both encoder and decoder have three different scales of temporal features.
 
Specifically, the encoder has two stages with the output from high resolution to low resolution, each stage is comprised of two convolution layers and a non-overlapping max-pooling layer. 

To further capture the global context in the lowest resolution, we propose to utilize a convolutional layer to bridge the feature from the encoder and decoder parts. 
For each convolutional layers mentioned above, we add a batch normalization~\cite{ioffe2015batch} and a ReLU activation function, \ie, 
\begin{equation}
\begin{aligned}
 \bm{M}^{e}_{2}= \mathtt{Pooling}(\mathtt{Conv}(\bm{M}^{e}_{1})), \\
 \bm{M}_{b}= \mathtt{Pooling}(\mathtt{Conv}(\bm{M}^{e}_{2})).
 \end{aligned}
\label{equation4}
\end{equation}
where $\bm{M}^{e}_{1}$, $\bm{M}^{e}_{2}$ represents the multi-modal features in encoder part, here $\bm{M}^{e}_{1}$ is the same value as $\bm{V}^{q}$. Here we use the depth-wise convolution for efficient network structure. More choice about the encoder layer is discussed in Section ~\ref{Computation Complexity}.

Then, the concentrated video feature in the lowest resolution (\ie, $\bm{M}_{b}$) are concatenated with language feature $\bm{\widetilde{Q}}$ again. To make the length of language and visual feature consistent, we use an weighted pooling on the language feature $\bm{\widetilde{Q}}$, \ie,
\begin{equation}
\begin{split}
 \bm{M}_{b} &= \mathtt{FFN}([\mathtt{Pooling}(\bm{W}_{m}\bm{M}_{b}),\mathtt{Pooling}(\bm{W}_{q}\bm{\bm{\widetilde{Q}}})]),
\label{equation}
\end{split}
\end{equation}

For the decoder, we try to refine the accurate boundaries of the responding region while recovering to high resolution. To be specific, the structure of the decoder part is symmetrical to the encoder part according to layer design. There are also two corresponding stages with the output from low resolution to high resolution in the decoder design. Each stage consists of two convolution layers, and is followed by a batch norm layer and a ReLU function, \ie, 
\begin{equation}
\begin{aligned}
 \bm{M}^{d}_{2}= \mathtt{Conv}([\mathtt{UpPooling}(\bm{M}_{b}),\bm{M}^{e}_{2}]), \\
 \bm{M}^{d}_{1}= \mathtt{Conv}([\mathtt{UpPooling}(\bm{M}^{d}_{2}),\bm{M}^{e}_{1}]).
 \end{aligned}
\label{equation}
\end{equation}
As shown in Figure~\ref{fig:architecture}, the input of each stage in the decoder part is the combination of output feature maps from the previous stage in the decoder part and those of the corresponding stage in the encoder part. In total, there are features in three different resolutions among the multi-resolution temporal modules. 
To fully supervise the generated temporal features in different resolutions, the output of each decoder block is supervised by the ground-truth temporal labels in their corresponding resolutions. To be specific, the ground-truth temporal labels can be transferred as 1D temporal maps $G$: The length of the whole video can be divided as $n$ clips (\eg, 128 clips), and all clips lie in the corresponding region of the query are represented as 1, the other clips are 0.

Based on the structure mentioned above, we add a $3\times 3$ convolution layer for each output feature map in the bridge stage and each decoder stage, followed by the upsampling and sigmoid functions.
Therefore, for input temporal features, our model produces three temporal attention maps of different resolutions in the training process. 
To obtain the 1D temporal maps of result, we use an 1D convolution layer with kernel of $3\times 3$ to output the features with one channel.
And we take $\bm{M}^{d}_{2}$ as the final input fed into the prediction module.

\subsection{Predictor Module}\label{Predictor}
We follow~\cite{zhang2020span} and construct a predictor by using two unidirectional LSTMs to predict the start and end logits according to feature $\bm{M}^{d}_{1}$. Meanwhile, two LSTMs are stacked so that prediction of end boundaries can be conditioned on those of start boundaries:
\begin{equation}
\begin{aligned}
     \mathbf{S}_{t}^{s} & = \mathbf{W}_{s}\times([\mathtt{UniLSTM}_\textrm{start}(\bm{M}_{1}^{d}, \bm{h}_{t-1}^{s});\bm{M}_{1}^{d}]) + \mathbf{b}_{s}, \\
    \mathbf{S}_{t}^{e} & = \mathbf{W}_{e}\times([\mathtt{UniLSTM}_\textrm{end}(\bm{S}_{t}^{s}, \bm{h}_{t-1}^{e});\bm{M}_{1}^{d}]) + \mathbf{b}_{e}.
\end{aligned}\label{eq_predictor}
\end{equation}
Here, $\bm{S}_{t}^{s}$ and $\bm{S}_{t}^{e}$ denote the scores of start and end boundaries at position $t$; $\bm{v}_{t}^{q}$ represents the $t$-th feature in $\bm{V}^{q}$.

\begin{table*}[!ht]
\caption{Statistics of Charades-STA, ActivityNet Caption, and TACoS datasets.}\label{table:dataset}
\small
\centering
\setlength{\tabcolsep}{2mm}
\begin{tabular}{|c|c|c|c|}
\hline

Dataset& Charades-STA& ActivityNet Captions& TACoS\\
\hline
\hline
Source &Homes & YouTube& Lab Kitchen\\
\hline
Domain& Indoor Activity & Open & Kitchen Cooking \\
\hline
\#Video & 6,672 &14,926 &1,27 \\
\hline
\#Annotation & 16,128 & 71,957& 18,818\\
\hline
\#Moments &11,767  & 71,957 & 3,290\\
\hline
Vocabulary Size &1,303  &15,505  &1,344 \\
\hline
Average Video Length (seconds) &30.60 &117.60 & 286.59\\
\hline
Average Moment Length (seconds) & 8.09 & 37.14 & 6.10\\
\hline
Average Query Length (words) & 7.22 & 14.41 & 10.05\\
\hline
Annotation Split (Training/ Validation/ Testing) &12,408 / - / 3,720 & 37,421/ 17,505 / 17,031 &10,146 / 4,589 / 4,083 \\
\hline
\end{tabular}
\end{table*}

Then, the probability distributions of start and end boundaries ($\bm{P}_{s} \in\mathbb{R}^{n}$, $\bm{P}_{e} \in\mathbb{R}^{n}$) are calculated by:
\begin{equation} \label{ps_pe}
\begin{aligned}
\bm{P}_{s} & =\mathtt{Softmax}(\bm{S}^{s}),   \\
\bm{P}_{e} & =\mathtt{Softmax}(\bm{S}^{e}),
\end{aligned}
\end{equation}

\textbf{Inference.} Suppose the video duration is $\mathcal{T}$, the start (end) index is calculated by $a^{s(e)}=\langle\tau^{s(e)}/\mathcal{T}\times n\rangle$, where $\langle\cdot\rangle$ denotes the rounding operator. During the inference, the predicted boundary can be easily converted to the corresponding time via $\hat{\tau}^{s(e)}= \hat{a}^{s(e)}/n\times \mathcal{T}$, and the predicted moment locations $(\hat{a}^{s},\hat{a}^{e})$ of a query is generated by maximizing the joint probability of start and end boundaries by:
\begin{equation}
\begin{aligned}
    \mathtt(\hat{a}^{s},\hat{a}^{e}) & = \arg\max_{\hat{a}^{s},\hat{a}^{e}} P_{s}(\hat{a}^{s}) P_{e}(\hat{a}^{e}) \\
    & \textrm{s.t. } 0\leq\hat{a}^{s}\leq \hat{a}^{e}\leq n
\end{aligned}
\end{equation}

\subsection{Loss function}\label{loss function}

To fully supervise the generated temporal features $M^{d}_{2}$, $M^{d}_{1}$, and $M_{b}$ in different resolutions, our training loss of multi-resolution temporal module is calculated by considering all outputs, \ie,
\begin{equation}
    L = {\sum}_{k=1}^K{\alpha_{k} l^{(k)}},
\end{equation}
where $l^{(k)}$ represents the loss of the $k$-th 1D temporal features, and $\alpha_{k}$ is the weight parameters. In our experiments, MRTNet is deeply supervised with outputs in three resolutions, \ie, $K=3$.

In each $l^{(k)}$, we define $l^{(k)}$ as a loss function to obtain high quality temporal location and clear boundaries:
\begin{equation}
    \centering
    L^{(k)} = L_{ce}^{(k)} + L_{ssim}^{(k)} + L_{iou}^{(k)},
    \label{equ:lside}
\end{equation}
where $l_{ce}^{(k)}$,~$l_{ssim}^{(k)}$, and $l_{iou}^{(k)}$ denote CE loss, SSIM loss, and IoU loss, respectively.

\textbf{CE loss} is widely used in previous video sentence grounding models:
\begin{equation}
    {L_{ce} = \frac{1}{2}\big[L_{\textrm{ce}}(P_{s}, Y_{s}) + L_{\textrm{ce}}(P_{e}, Y_{e})\big]},
\end{equation}
where $L_{\textrm{CE}}$ represents cross-entropy loss function; $Y_{s}$ and $Y_{e}$ are the labels for the start ($a^{s}$) and end ($a^{e}$) boundaries, respectively. $P_{s}$ and $P_{e}$ are generated according to Eq.~\ref{ps_pe}.

\textbf{SSIM (Structural Similarity Index Measure) loss}~\cite{wang2003multiscale}, is first proposed in image quality assessment task. Since there exists structural information in an image, SSIM can effectively measure the similarity of images, which has been proven in relevant tasks~\cite{qin2019basnet}. In the video moment grounding task, 1D temporal structural information is also important, which can be utilized as a training loss compared with ground truth. 

For two clips from prediction $S$ and the binary 1D temporal groundtruth $G$, we calculate the SSIM loss by elements in each clip. We denote $x=\{x_j | j=1,...,N\}$ and $y=\{y_j | j=1,...,N\}$, $x$ is cropped from $S$ and $y$ is cropped from $G$. 
respectively. Then, the SSIM loss of $x$ and $y$ can be calculated as introduced below:
\begin{equation}
    \centering
    {L_{ssim}=1 -  \frac{(2\mu_x\mu_y+C_1)(2\sigma_{xy}+C_2)}{(\mu_x^2+\mu_y^2+C_1)(\sigma_x^2+\sigma_y^2+C_2)}}
    \label{equ:ssim_loss}
\end{equation}
Here $\mu_x$, $\mu_y$ represents the mean deviations of $x$ and $y$, $\sigma_x$, $\sigma_y$ represents the standard deviations of $x$ and $y$. $\sigma_{xy}$ represents covariance. $C_1$ and $C_2$ are parameters that are introduced in implementation details. 

\textbf{IoU loss} is widely used in segmentation tasks, and can also be borrowed into the video sentence grounding task. The calculation is listed below:
\begin{equation}
    \centering
    L_{iou} = 1 - \tfrac{\sum\limits_{c=1}\limits^WS(c)G(c)}{\sum\limits_{c=1}\limits^W[S(c)+G(c)-S(c)G(c)]},
    \label{equ:liou}
\end{equation}
Here $G(c)\in\{0,1\}$ denotes the groundtruth label of the element $c$ and $S(c)$ is the prediction of 1D temporal map.

\section{Experiments}

\subsection{Datasets}
To evaluate the performance of our proposed MRTNet, we conduct experiments on three challenging video sentence grounding datasets:

\begin{itemize}
    \item \textbf{Charades-STA}~\cite{gao2017tall} is composed of daily indoor activities videos, which is based on Charades dataset~\cite{sigurdsson2016hollywood}. This dataset contains 6672 videos, 16,128 annotations, and 11,767 moments. The average length of each video is 30 seconds. $12,408$ and $3,720$ moment annotations are labeled for training and testing, respectively;
    
    \item \textbf{ActivityNet Caption}~\cite{caba2015activitynet} is originally constructed for dense video captioning, which contains about $20$k YouTube videos with an average length of 120 seconds. As a dual task of dense video captioning, video sentence grounding utilize the the sentence description as query and output the temporal boundary of each sentence description.  
    
    \item \textbf{TACoS}~\cite{regneri2013grounding} is collected from MPII Cooking dataset~\cite{regneri2013grounding}, which has 127 videos with an average length of $286.59$ seconds. 
    TACoS has 18,818 query-moment pairs, which are all about cooking scenes.
    We follow the same splits in~\cite{gao2017tall}, where $10,146$, $4,589$ and $4,083$ annotations are used for training, validation and test, respectively.
\end{itemize}

The details of each dataset are summarized in Table~\ref{table:dataset}. It it worth noting that some methods make minor changes to the dataset when evaluating the experimental performance. For example, CMIN~\cite{zhang2019cross} uses val\_1 as validation set and val\_2 as testing set in the ActivityNet Captions dataset, while other methods~\cite{zhang2020learning} combine the val\_1 and val\_2 together as the testing set. And for TACoS dataset, 2D-TAN~\cite{zhang2020learning} utilize a modified TACoS dataset for evaluation. To make a fair comparison, we follow the setting of dataset splitting report in their original papers when evaluating the performance of our method.

\subsection{Evaluation Metrics}
Following existing video grounding works, we evaluate the performance on two main metrics:

\textbf{ mIoU:}
``mIoU" is the average predicted Intersection over Union over all testing samples. The mIoU metric is particularly challenging for short video moments;

\textbf{ Recall:}
We adopt ``$\textrm{R@}n, \textrm{IoU}=\mu$'' as the evaluation metrics, following~\cite{gao2017tall}. The ``$\textrm{R@}n, \textrm{IoU}=\mu$'' represents the percentage of language queries having at least one result whose IoU between top-$n$ predictions with ground-truth is larger than $\mu$.  In our experiments, we reported the results of $n=1$ and $\mu\in\{0.3, 0.5, 0.7\}$.

\subsection{Implementation Details}\label{implementation}
For language query $Q$, we use the $300$-D GloVe~\cite{pennington2014glove} vectors to initialize each lowercase word, and these word embeddings are fixed during training.
For video $V$, we downsample frames and extracted RGB visual features using the 3D ConvNet which was pre-trained on the Kinetics dataset. We set the dimension of all the hidden layers in the model as $128$, the kernel size of the convolutional layer as $7$, and the head size of multi-head attention as $8$. For all datasets, models were trained for $100$ epochs The batch size was set to $16$. Dropout and an early stopping strategies were adopted to prevent overfitting. The whole framework was trained by Adam optimizer with an initial learning rate 0.0001. $C_1=0.01^2$ and $C_2=0.03^2$.

\begin{table}[!t]
 \caption{Performance comparison (\%) with the state-of-the-art models on TACoS dataset. Noted that the improvement achieved by MRTNet is significant.} \label{experiment_tacos}
    \centering
    \small
     \setlength{\tabcolsep}{1.1mm}{
        \begin{tabular}{l|c|c|c|c}
        \hline
        Methods  &\textrm{R@}1,IoU=0.3 & \textrm{R@}1,IoU=0.5 & \textrm{R@}1,IoU=0.7 & mIoU  \\ \hline
 
        CTRL \tiny ICCV2017    & 18.32   & 13.30     & --      & --    \\
        ACRN \tiny SIGIR 2018    & 19.52   & 14.62      & --      & --    \\
        MCF \tiny IJCAI2018     & 18.64   & --      & --      & --    \\
        SM-RL \tiny CVPR2019    & 20.25   & 15.95      & --      & --    \\
        ACL \tiny WACV2019     & 22.07   & 17.78      & --      & --    \\
        SAP \tiny AAAI2019     & --      & 18.24      & --      & --    \\
        L-NET \tiny AAAI2019    & --      & --      & --      & 13.41 \\
        TGN \tiny EMNLP2018     & 21.77    & 18.90   &  --      & -- \\
        ABLR-aw \tiny AAAI2019  & 18.90    & 9.30      & --      & 12.50  \\
        ABLR-af \tiny AAAI2019  & 19.60    & --      & --      & 13.40  \\
        DEBUG \tiny EMNLP2019   & 23.45   & 11.72      & --      & 16.03 \\
        CMIN \tiny SIGIR2019  & 24.64  & 18.05      & --      & -- \\
        GDP \tiny AAAI2020   & 24.14   & --          & --      & 16.18 \\
        DRN \tiny CVPR2020 &-- &23.17 &--  &--\\
        SeqPAN \tiny ACL2021 & 31.72 & 27.19 & 21.65 & 25.86 \\
        \hline
        \hline
        VSLNet \tiny ACL2020  & 29.61   & 24.27   & 20.03   & 24.11 \\
   
        MRTNet (VSLNet)     & \textbf{32.35}   & \textbf{25.84}   & \textbf{21.31}   & \textbf{26.14} \\ 
        \hline
        \hline
         2D-TAN \tiny AAAI2020       & 37.29    & 25.32  & 13.32 & 25.19  \\
     MRTNet (2D-TAN)         & \textbf{37.81}  & \textbf{26.01}   & \textbf{14.95}   & \textbf{26.29}\\ 
        
       \hline 
        \end{tabular}}
\end{table}
  
\begin{table}[!h]
\caption{Performance comparison (\%) with the state-of-the-art models on ActivityNet Captions dataset. Noted that the improvement achieved by MRTNet is significant.} \label{performance_activitynet}
    \centering
    \small
    \setlength{\tabcolsep}{1.1mm}{
    \begin{tabular}{l|c|c|c|c}
    \hline
    Methods  & \textrm{R@}1,IoU=0.3 & \textrm{R@}1,IoU=0.5 & \textrm{R@}1,IoU=0.7 & mIoU  \\ \hline
    CMIN \tiny SIGIR2019  &63.61    & 43.40   & 23.88      & --    \\
    TGN \tiny EMNLP2018         & 43.81   & 27.93   & --      & --    \\
    
    QSPN \tiny AAAI2019        & 45.30    & 27.70    & 13.60      & --    \\
    RWM \tiny AAAI2019        & --      & 36.90    & --      & --    \\
    ABLR-af \tiny AAAI2019    & 53.65   & 34.91   & --      & 35.72 \\
    ABLR-aw \tiny AAAI2019    & 55.67   & 36.79   & --      & 36.99 \\
    DEBUG \tiny EMNLP2019      & 55.91   & 39.72   & --      & 39.51 \\
    GDP \tiny AAAI2020        & 56.17   & 39.27   & --       & 39.80 \\
    DRN \tiny CVPR2020         & 58.52 &41.51 &23.07 & 41.13  \\
    CI-MHA \tiny SIGIR2021  & 61.49 &43.97  &25.13  & -- \\
    SeqPAN \tiny ACL2021       & 61.65 & 45.50 & 29.37 & 45.11 \\
    \hline
    \hline
    VSLNet \tiny ACL2020     & 63.16   & 43.22   & 26.16   & 43.19 \\
   
    MRTNet (VSLNet)        & \textbf{64.17}   & \textbf{44.09}   & \textbf{27.43}   & \textbf{44.82} \\ 
  
    \hline
    \hline

    2D-TAN \tiny AAAI2020       &  59.45   & 44.51  & 27.38   & 43.29 \\
     MRTNet (2D-TAN)         & \textbf{60.71}  & \textbf{45.59}   & \textbf{28.07}   & \textbf{44.54}\\ 
     \hline
    \end{tabular}}
\end{table}

\begin{table}[!h]
    \caption{Performance (\%) comparison with the state-of-the-art models on Charades-STA dataset. Noted that the improvement achieved by MRTNet is significant.}\label{performance_charades} 
    \centering
    \small
    \setlength{\tabcolsep}{0.9mm}{
    \begin{tabular}{l|c|c|c|c}
    \hline
    Methods  & \textrm{R@}1,IoU=0.3 & \textrm{R@}1,IoU=0.5 & \textrm{R@}1,IoU=0.7 & mIoU  \\ \hline
     CTRL \tiny ICCV2017        & --      & 23.63   & 8.89    & --    \\
 
    ROLE \tiny MM2018         & 25.26   & 12.12   & --      & --    \\
    ACL \tiny WACV2019       & --   & 30.48   & 12.20   & --    \\
   SAP \tiny AAAI2019          & --      & 27.42   & 13.36   & --    \\
    RWM \tiny AAAI2019         & --      & 36.70    & --      & --    \\
  
    SM-RL \tiny CVPR2019        & --      & 24.36   & 11.17   & --    \\
    QSPN \tiny AAAI2019    & 54.70    & 35.60    & 15.80    & --    \\
    DEBUG \tiny EMNLP2019       & 54.95   & 37.39   & 17.92   & 36.34    \\
    GDP \tiny AAAI2020          & 54.54   & 39.47   & 18.49   & --     \\
    DRN \tiny CVPR2020 &- &42.90 &23.68 &41.28 \\ 
    
    ExCL \tiny ACL2019    & --      & 44.10    & 22.40    & --    \\
    MAN \tiny CVPR2019         & --        & 46.53   & 22.72    & --       \\
    CI-MHA \tiny SIGIR2021 & 69.87        &54.68   & 35.27    & --       \\
    \hline    
    \hline
    VSLNet \tiny ACL2020       & 70.46    & 54.19   & 35.22   & 50.02 \\

    MRTNet (VSLNet)         & \textbf{70.88}  & \textbf{56.19}   & \textbf{36.37}   & \textbf{50.74}\\ 
    \hline
    \hline
 
    2D-TAN \tiny AAAI2020       &  57.31   & 42.80  & 23.25   & 39.23 \\
     MRTNet (2D-TAN)         &  \textbf{59.23}  & \textbf{44.27}   & \textbf{25.88}   &  \textbf{40.59}\\ 
     \hline
    \end{tabular}}
\end{table}

\subsection{Comparison with State-of-the-Arts}

\subsubsection{Experimental Settings.}
We compare our proposed MRTNet with state-of-the-art video sentence grounding methods on three public datasets. These methods can be grouped into three categories according to the viewpoints of proposal-based, proposal-free, and other approaches:

1) Proposal-based models: 
 \textbf{CTRL}~\cite{gao2017tall},
 \textbf{ACRN}~\cite{liu2018attentive}, 
 \textbf{ROLE}~\cite{liu2018cross}, 
 \textbf{MCF}~\cite{wu2018multi}, \textbf{ACL}~\cite{ge2019mac}, 
 \textbf{SAP}~\cite{chen2019semantic},
\textbf{QSPN}~\cite{xu2019multilevel},
\textbf{CMIN}~\cite{zhang2019cross}
\textbf{TGN}~\cite{chen2018temporally},
\textbf{2D-TAN}~\cite{zhang2020learning}
\textbf{MAN}~\cite{zhang2019man};
2) Proposal-free models:
\textbf{L-Net}~\cite{chen2019localizing}, 
\textbf{ABLR-aw}~\cite{yuan2019find},
\textbf{ABLR-af}~\cite{yuan2019find},
\textbf{DEBUG}~\cite{lu2019debug},
\textbf{CI-MHA}~\cite{yu2021cross}
\textbf{ExCL}~\cite{ghosh2019excl},
\textbf{GDP}~\cite{chen2020rethinking}, \textbf{VSLNet}~\cite{zhang2020span},
\textbf{LGI}~\cite{mun2020local},
\textbf{DRN}~\cite{zeng2020dense}
\textbf{SeqPAN}~\cite{zhang2021parallel};
3) Others:
\textbf{RWM}~\cite{he2019read},
\textbf{SM-RL}~\cite{wang2019language}.

\begin{figure*}[!t]
    \centering
   \includegraphics[width=\linewidth]{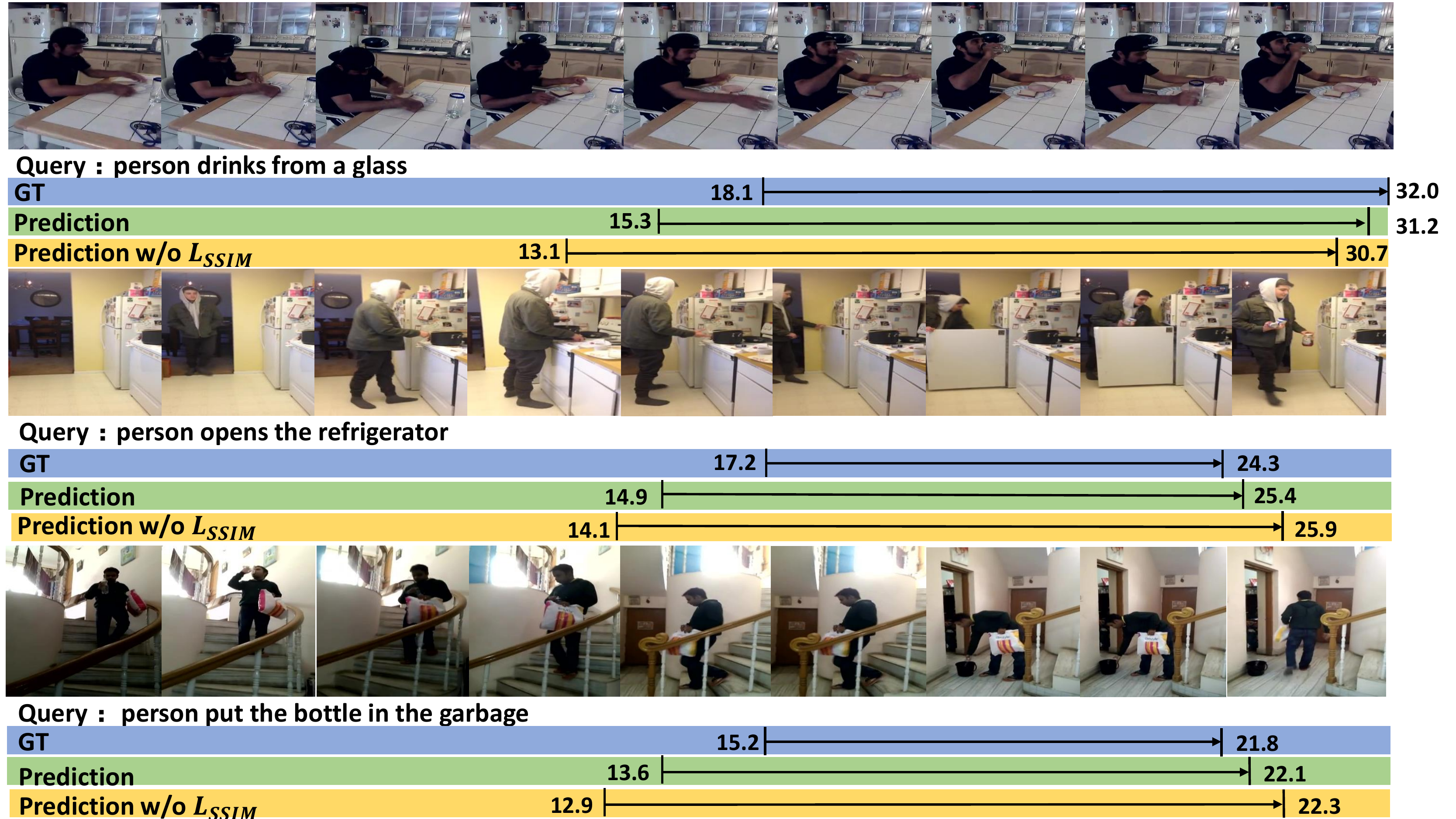}
    \caption{Qualitative results of MRTNet (VSLNet) on Charades-STA dataset. }
    \label{fig:experiment}
\end{figure*}

\subsubsection{Quantitative Results.}

The results on three benchmarks are reported from Table ~\ref{experiment_tacos} to Table ~\ref{performance_charades}, respectively. According to the experimental results, we can observe that our MRTNet can effectively improve the performance of baseline networks over all metrics and benchmarks. We choose VSLNet~\cite{zhang2020span} and 2D-TAN~\cite{zhang2020learning} as baseline networks, which are known as typical proposal-free and proposal-based models with published source codes. 

\begin{itemize}
    \item For the implementation based on each baseline method, our MRTNet (VSLNet) shares the same architecture as VSLNet in Feature Encoder Module. Specifically, we implement the VSLNet with C3D feature followed the settings they reported in original paper. VSLNet extracts the frame-level feature using the backbone of QANet~\cite{yu2018qanet} and utilizes two LSTMs to classify the start and end boundary. 
    For the MRTNet (2D-TAN), we add the MRT Module before restructuring the 2D temporal feature map.  For other modules in each baseline network, we follow the same settings according to original papers.
    
    \item Table~\ref{performance_charades} summarizes the experimental results on Charades-STA dataset. We can observe that MRTNet (VSLNet) works well in even stricter metrics, \eg, MRTNet (VSLNet) achieves a significant 1.28\% absolute improvement in IoU@0.7 and MRTNet (2D-TAN) achieves 1.63\% absolute improvement in IoU@0.7, which demonstrates the effectiveness of proposed model. It is mainly because that MRTNet better utilizes the temporal consistency of video sequences in different resolutions and utilizes a loss function to better regularize the multi-modal features. The results on TACoS and ActivityNet Captions dataset are summarized in Table~\ref{experiment_tacos} and Table~\ref{performance_activitynet}, respectively. Note that videos in TACoS have a longer average duration and the ground-truth video segments in ActivityNet Captions have a longer average length. 
    Although there exists bias in the data distribution of datasets~\cite{yang2021deconfounded}, MRTNet significantly outperforms the other methods on both benchmarks with the C3D feature, which demonstrates that MRTNet is highly adaptive to videos and segments with diverse lengths, and can bring in consistent improvement based on different baseline methods.
    
     \item Our proposed MRT module also has good generalization ability, since the input multi-modal features of the MRT module keep in the same dimension as output features, then this module can be easily plugged into any other proposal-based or proposal-free video grounding network, and the whole network with proposed loss function can be trained in an end-to-end manner.
  
\end{itemize}

\begin{figure*}[t]
\centering
{
\includegraphics[width=0.32\textwidth]{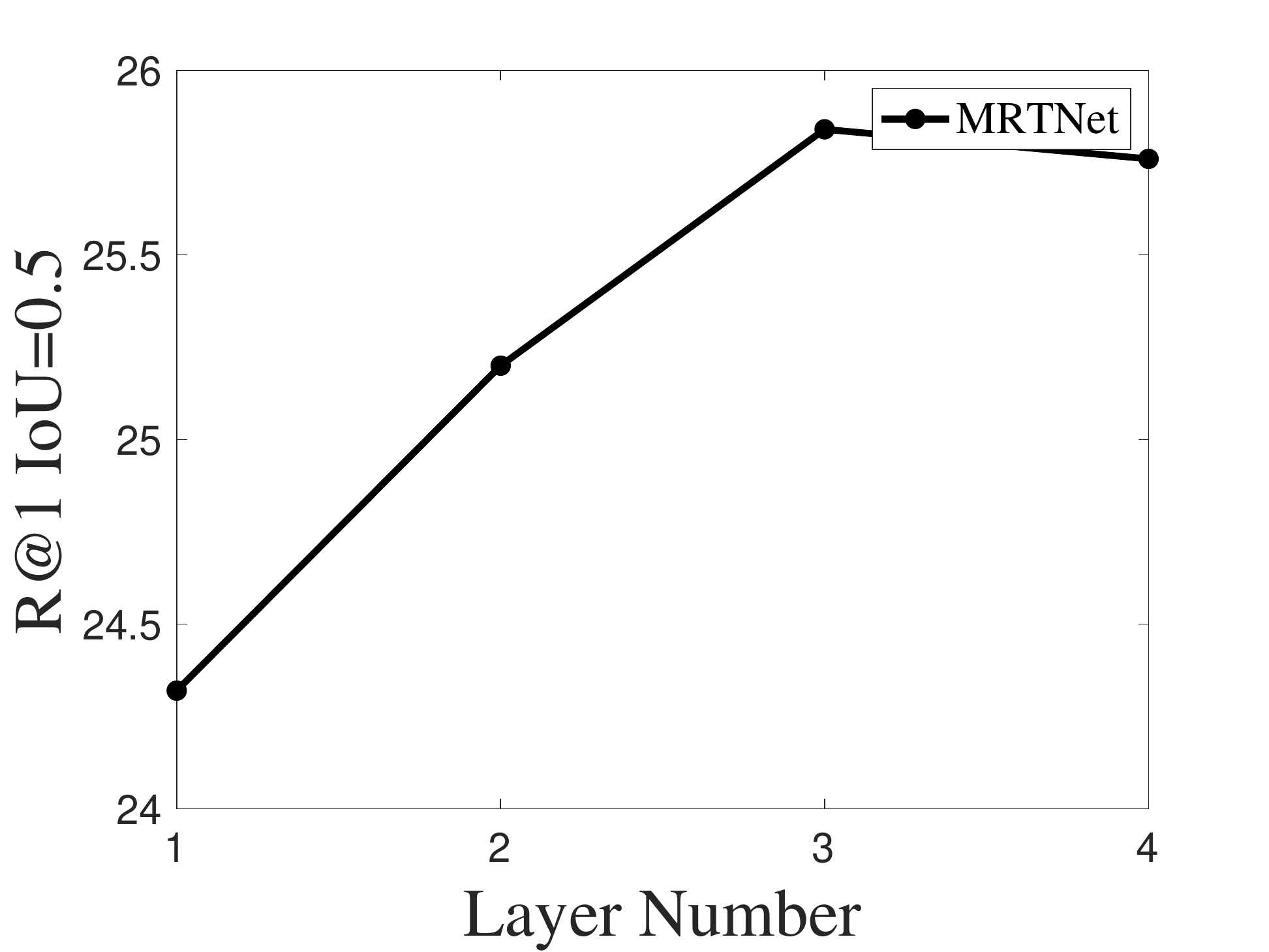}
\includegraphics[width=0.32\textwidth]{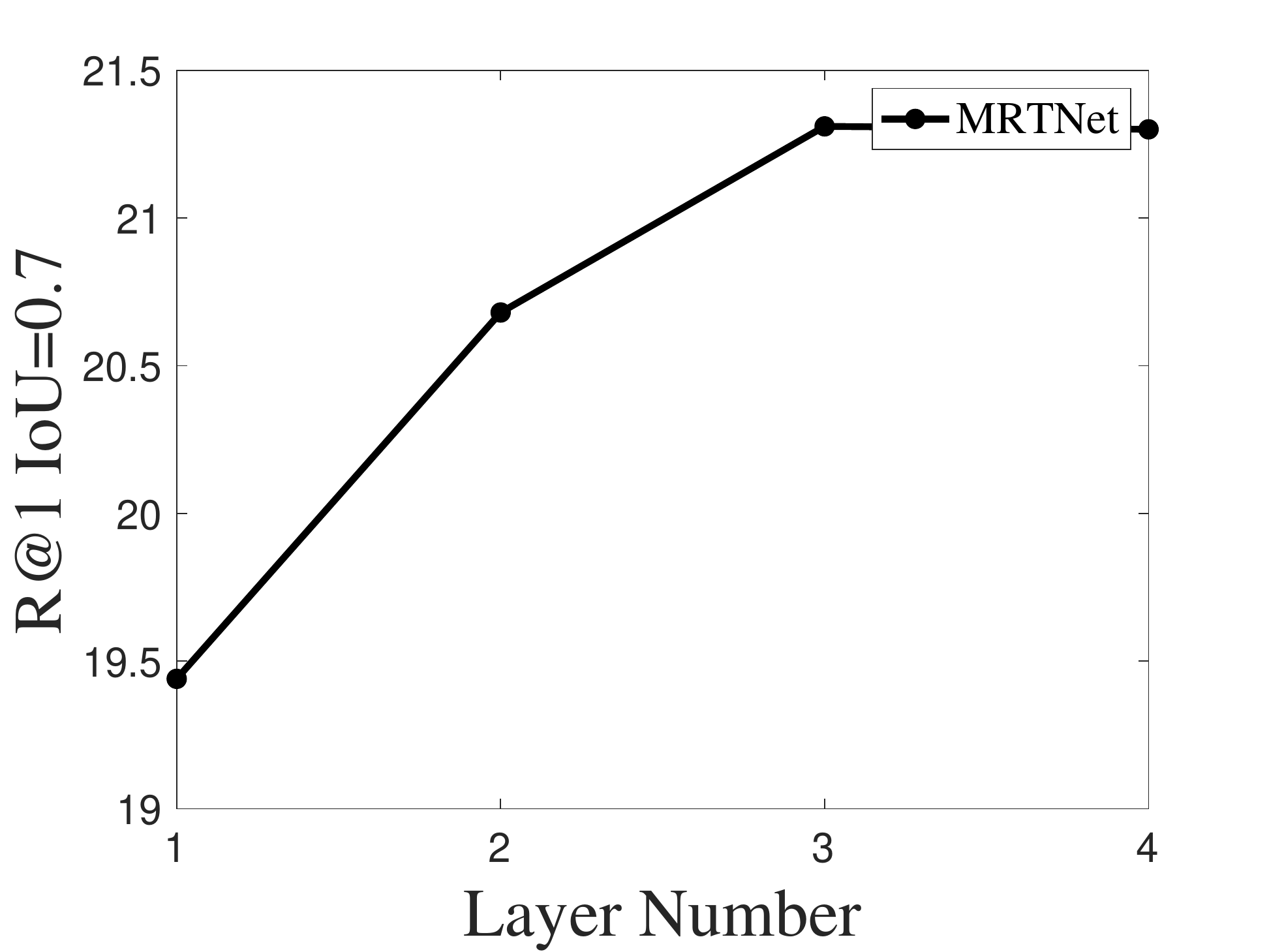}
\includegraphics[width=0.32\textwidth]{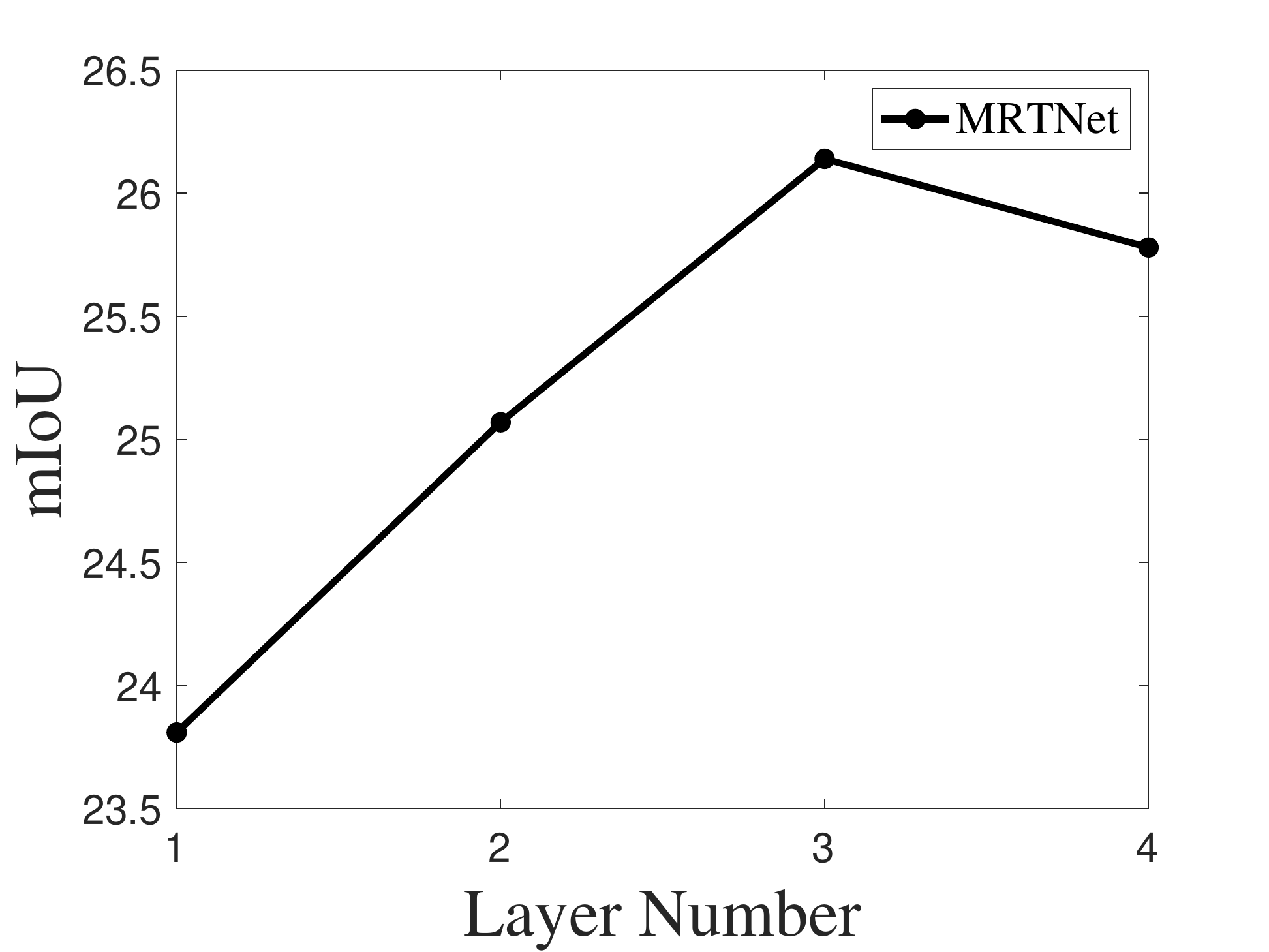}
\caption{Performance comparison (\%) of MRTNet (VSLNet) in different layer numbers on TACoS dataset.}\label{layer_number}
}
\end{figure*}

\subsubsection{Qualitative Results.}

As illustrated in Figure~\ref{fig:experiment}, the qualitative results of MRTNet (VSLNet) on Charades-STA dataset are reported. According to the three examples, the localized moments generated by MRTNet are very  close to ground-truth. If we train the model without SSIM loss, the generated temporal region will be more expansive compared with the ground-truth region, which is shown as a longer temporal interval with high confidence. Hence, the model trained with SSIM loss can better capture the accurate boundaries of query-corresponding temporal regions.

\subsection{Ablative Studies}\label{ablation}

We conduct ablative experiments to analyze the effectiveness of loss function in our approach. Also, we analyze the effectiveness of the MRT module in MRTNet.
For simplification, the ablation experiments introduced below are conducted on the TACoS dataset with VSLNet as the baseline. 

\subsubsection{Multi-resolution Temporal Module.}

From the results in Figure~\ref{layer_number} below, we can observe that three layers in the encoder-decoder structure is the most appropriate setting when dealing with videos of different lengths, which can achieve a trade-off between speed and accuracy. 

\begin{table}[!ht]
\caption{Performance comparison (\%) of MRTNet (VSLNet) with different encoding layers on TACoS dataset.}\label{conv_vs_transformer}

\centering
\setlength{\tabcolsep}{2mm}
\begin{tabular}{|c|c|c|c|}
\hline
Encoding Layer& \textrm{R@}1, IoU=0.5& \textrm{R@}1, IoU=0.7& mIoU\\
\hline
\hline
Convolution&25.84 & 21.31& 26.14\\
\hline
Transformer&25.91 &21.14 &26.21 \\

\hline
\end{tabular}
\end{table}

Then, we consider replacing the convolution layer with vanilla Transformer in the encoder part of multi-resolution module, as shown in Eq.~\eqref{equation4}. According to Table~\ref{conv_vs_transformer}, we can observe that the performance of using Transformer layer in the encoder part is comparative with using convolution, but causes larger computation cost. To make a trade-off between speed and accuracy, our MRTNet use depth-wise convolution in the encoder layer. 

\begin{table}[htp]
\caption{Performance comparison (\%) of MRTNet (VSLNet) with different losses on TACoS dataset. }

\centering
\setlength{\tabcolsep}{1.6mm}{
\begin{tabular}{|c|ccc|c|c|}
\hline

\multirow{1}{*}{Loss} &
 
  \multicolumn{1}{c|}{$\ell_{ce}$} &
  \multicolumn{1}{c|}{$\ell_{ssim}$} &
  \multicolumn{1}{c|}{$\ell_{iou}$} & 
  \multicolumn{1}{c|}{\textrm{R@}1, IoU=0.7} & 
    \multicolumn{1}{c|}{mIoU}

   \\ \hline
   \hline
\multirow{4}{*}{\begin{tabular}[c]{@{}c@{}}MRT Module \end{tabular}} &
  \multicolumn{1}{c|}{\checkmark} &
  \multicolumn{1}{c|}{} &
  \multicolumn{1}{c|}{}& 
   20.01&
  24.94  \\
  
 &
  \multicolumn{1}{c|}{\checkmark} &
  \multicolumn{1}{c|}{\checkmark} &
  \multicolumn{1}{c|}{}&
  20.30&
  25.26 \\
 &

  \multicolumn{1}{c|}{\checkmark} &
  \multicolumn{1}{c|}{} &
  \multicolumn{1}{c|}{\checkmark}&
  20.24&
  25.29 \\
 &
  \multicolumn{1}{c|}{\checkmark} &
  \multicolumn{1}{c|}{\checkmark} &
  \multicolumn{1}{c|}{\checkmark} &
  21.31&
  26.14 \\ \hline

\end{tabular}}

\label{hybrid}
\end{table}

\subsubsection{Loss function.} 
To evaluate the effectiveness of each component in our proposed loss function, we conduct a series of experiments with different losses:
\begin{itemize}
    \item The CE loss is computed in the frame-level. It mainly evaluates the possibility of two peaks in time dimension: start and end timestamps. As shown in Table~\ref{hybrid}, training the MRTNet (VSLNet) model with merely CE loss achieves 24.94 in mIoU.
    
    \item The SSIM loss is a clip-level measure. The results in Table~\ref{hybrid} indicate that the proposed SSIM loss greatly improves the performance, which brings about 0.7\% performance gain on average in the metric of mIoU. The SSIM loss can regularize the features at clip-level and refine the temporal boundaries.

    \item  The IoU loss is a sequence-level measure. 
As shown in Table~\ref{hybrid}, utilizing SSIM loss can bring about 0.3\% performance gain in the metric of mIoU.
\end{itemize}

According to the detailed analysis of each loss, it is clear that our loss function is beneficial to better feature learning and can achieve superior qualitative results.

\subsubsection{Computation Complexity.}\label{Computation Complexity}
We evaluate the computation complexity of MRTNet compared with other SOTA methods on TACoS dataset. 
\begin{itemize}
    \item According to Table ~\ref{experiment_tacos} and Table ~\ref{Computation Complexity}, MRTNet achieves better performance compared with baseline methods in all experimental cases. Due to the utilization of depth-wise convolution, the model size of whole network has only minor increment (almost 6$\%$) compared with the baseline 2D-TAN method.  
    \item We also try to replace the convolution layer in encoder part of MRT Module with vanilla Transformer layer, which can further improve the performance but brings in more parameters and increase the running time. 
    \item The experimental results show that utilizing the characteristic of multi-resolution in video can bring in steady improvement based on each baseline network and still have potential for better performance. To be specific, we can use light-weighted version convolution or Transformer layers to achieve better performance while relieving the burden of computation cost. Also, there also exists a series of variants designed for multi-scale feature interaction, such as pyramid pooling, multi-scale attention, and so on.
\end{itemize}

\begin{table}[!ht]
\caption{Computation Complexity Comparison with SOTA methods on TACoS dataset.}\label{Computation Complexity}
\small
\centering
\setlength{\tabcolsep}{2mm}
\begin{tabular}{|c|c|c|c|}
\hline
Method& Running Time& Model Size& \textrm{R@}1, IoU=0.5\\
\hline
\hline
CTRL & 2.23s & 22M & 13.30\\
ACRN & 4.31s & 128M & 14.62\\
TGN  &0.92s & 166M & 18.90\\
DRN &0.15s &214M & 23.17 \\
2D-TAN &0.57s &232M & 25.32 \\
\hline
MRTNet (2D-TAN) &0.68s & 249M & 26.01 \\
+Transformer& 0.82s & 284M &26.21 \\

\hline
\end{tabular}
\end{table}

\section{Conclusion}
In this work, we propose a novel multi-resolution temporal video sentence grounding network: MRTNet, which mainly consists of a multi-model feature encoder, a Multi-Resolution Temporal (MRT) module, and a predictor. 
Our MRT module is plug-and-play, which means it can be seamlessly incorporated into any video sentence grounding baseline network. Besides, we propose a loss function to supervise the cross-modal features of accurate video sentence grounding in three levels: frame-level, clip-level, and sequence-level. Through extensive experiments on three benchmark datasets, we show that  MRTNet outperforms the baseline methods; and the proposed MRT module is easy to plug into other video sentence grounding frameworks. 
In future work, we will explore more efficient and light-weighted models to make quick video sentence grounding inference.

{\small
\bibliographystyle{ACM-Reference-Format}
\bibliography{mrtnet}
}


\end{document}